\documentclass[letterpaper]{article} 
\usepackage{aaai2027-xetex-draft}  
\usepackage[hyphens]{url}  
\usepackage{graphicx} 
\usepackage{multirow}
\usepackage{booktabs}
\usepackage{amsmath}
\usepackage{algorithm}
\usepackage{algpseudocode}

\usepackage{natbib}  
\usepackage{caption} 
\AtBeginDocument{}

\providecommand{\pdfinfo}[1]{}
\title{UAV-OVVIS: Unmanned Aerial Vehicles Also Need Open-Vocabulary Video Instance Segmentation}
\author{
    Mingyu Dou\textsuperscript{\rm 1,2},
    Shi Qiu\textsuperscript{\rm 1,2}\thanks{Corresponding author. Email: qiushi@opt.ac.cn},
    Ming Hu\textsuperscript{\rm 1,2},
    Yifan Chen\textsuperscript{\rm 3,4},
    Zhe Sun\textsuperscript{\rm 3,5}\thanks{Corresponding author. Email: sunzhe@nwpu.edu.cn}
}
\affiliations{
    \textsuperscript{\rm 1}Xi'an Institute of Optics and Precision Mechanics, Chinese Academy of Sciences\\
    \textsuperscript{\rm 2}University of Chinese Academy of Sciences\\
    \textsuperscript{\rm 3}Institute of Artificial Intelligence (TeleAI), China Telecom\\
    \textsuperscript{\rm 4}College of Future Information Technology, Fudan University\\
    \textsuperscript{\rm 5}School of Artificial Intelligence, Optics and Electronics, Northwestern Polytechnical University\\
}
\nocopyright

\begin{document}

\maketitle

\begin{abstract}
Unmanned Aerial Vehicle (UAV) videos are widely used in traffic monitoring, urban management, and emergency rescue. However, existing UAV video perception mainly relies on box-level localization and trajectory association under predefined categories, making it difficult to simultaneously support flexible queries and fine-grained instance-level dynamic understanding in open scenarios. To this end, we introduce a new task, UAV Open-Vocabulary Video Instance Segmentation (UAV-OVVIS), which discovers targets in UAV videos according to open-vocabulary queries and outputs instance-level segmentation trajectories with globally consistent identities. Considering the scarcity of instance-level annotations in UAV scenarios, we propose AeroTrack, a training-free unified framework. AeroTrack centers on periodic open-vocabulary detection, short-segment mask propagation, and cross-segment identity unification, reusing existing visual foundation models to enable UAV-OVVIS. Based on this framework, we instantiate five AeroTrack variants and construct AeroVIS, an evaluation benchmark for UAV-OVVIS containing 9 UAV object categories and 8,279 trajectories. Experiments show that AeroTrack substantially outperforms existing general video instance segmentation methods in UAV scenarios and demonstrates strong open-vocabulary robustness and generalization. To support future research, we release AeroTrack and AeroVIS as a unified framework and benchmark for UAV-OVVIS.
\end{abstract}

\begin{links}
    \link{Code}{https://github.com/Dmygithub/AeroTrack}
\end{links}

\begin{figure*}[t]
\centering
\includegraphics[width=0.98\textwidth]{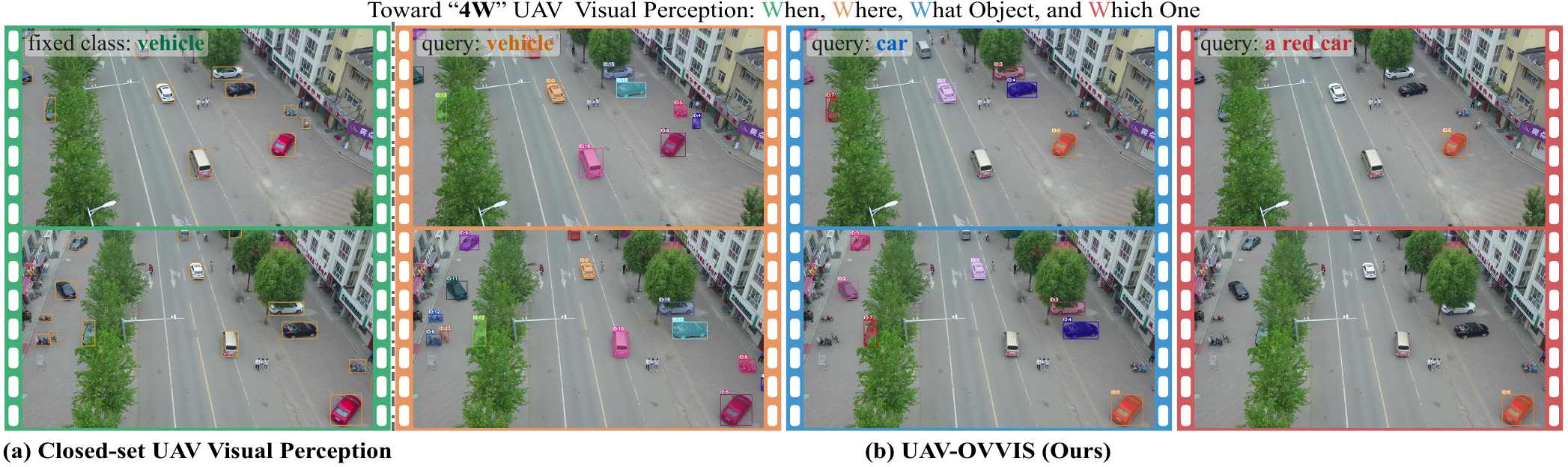}
\caption{(a) \textbf{Closed-set UAV Visual Perception}, represented by UAV multi-object tracking, such as box-level trajectories for a \textit{fixed class: vehicle}. (b) \textbf{UAV-OVVIS (Ours)} targets \textbf{4W} perception, namely When (temporal perception), Where (fine-grained localization), What Object (open-vocabulary semantics), and Which One (instance identity). It supports open-vocabulary queries at different granularities, such as \textit{vehicle}, \textit{car}, and \textit{a red car}, and outputs instance-level segmentation trajectories with globally consistent video identities.}
\label{fig:motivation}
\end{figure*}

\section{Introduction}

Unmanned Aerial Vehicle (UAV) videos provide continuous and dynamic visual perception over large spatial areas, and play an important role in applications such as traffic monitoring, urban management, and emergency rescue~\cite{wen2021dronecrowd}. Existing UAV video perception has long relied on box-level localization and trajectory association under predefined categories. Related detection, recognition, and tracking methods and benchmarks have formed a relatively mature benchmark ecosystem~\cite{li2025remdet,ma2026amot}, establishing the basis of visual perception from the UAV perspective. Meanwhile, recent studies on open-category video instance segmentation indicate that visual perception is moving from closed-set categories toward open-vocabulary semantics~\cite{guo2025openvis}. With advances in visual foundation models (VFMs), UAV video perception is also extending toward open-vocabulary understanding and fine-grained instance-level perception. Existing studies have explored open-vocabulary object discovery and tracking in UAV videos, allowing target semantics to be flexibly specified by text queries at inference time~\cite{blei2025cloudtrack,chen2026aerialmind}. At the same time, Video Instance Segmentation (VIS) has gradually been introduced into UAV scenarios~\cite{huang2025uavvis}. By jointly modeling detection, segmentation, and instance association under predefined categories, VIS outputs instance-level segmentation trajectories and further improves fine-grained understanding of dynamic UAV scenes. However, existing UAV dynamic perception paradigms still struggle to simultaneously provide the flexibility of open-vocabulary querying and the precision of instance-level segmentation trajectories, limiting their ability to meet the demands of fine-grained dynamic understanding in open scenarios.

To address the limitations of existing UAV perception paradigms, we introduce \textbf{UAV Open-Vocabulary Video Instance Segmentation (UAV-OVVIS)}, which discovers specified targets in UAV videos according to open-vocabulary text queries and outputs instance-level segmentation trajectories with globally consistent video identities. As shown in Figure~\ref{fig:motivation}, UAV-OVVIS is designed for \textbf{4W} UAV Visual Perception. (a) Traditional \textit{Closed-set UAV Visual Perception}, exemplified by UAV multi-object tracking, can only perceive predefined categories such as \textit{vehicle} and produce box-level tracking trajectories. (b) In contrast, \textit{UAV-OVVIS (Ours)} supports open-vocabulary queries at different semantic granularities and outputs instance-level segmentation trajectories with globally consistent video identities.

Given the scarcity of instance-level annotations in UAV scenarios, and the strong general capabilities that large-scale pretrained visual foundation models (VFMs) have demonstrated in open-vocabulary detection, instance-level segmentation, and mask propagation, this paper focuses on enabling UAV-OVVIS by reusing existing VFMs in a training-free manner. However, this approach still faces several challenges. First, UAV-OVVIS needs to be decoupled into coordinated foundation vision tasks, where multiple VFMs collaborate to perform open-vocabulary detection, instance-level segmentation and propagation, and global identity maintenance. Second, the joint motion of UAV platforms and targets, small and densely distributed objects, occlusion, and frequent entry and exit from the field of view require the system to continuously cover text-specified targets in long videos, while distinguishing newly entered instances from reappearing historical ones. Finally, open-vocabulary detection, instance-level segmentation, and temporal propagation substantially amplify memory and computational costs, making per-frame detection or whole-video propagation difficult to satisfy practical UAV video processing requirements.

To address these challenges, we propose a unified \textbf{AeroTrack} framework that decouples UAV-OVVIS into two components: an Open-Vocabulary Recognizer (Recognizer) and a Video Instance Segmenter (Segmenter). Specifically, the Recognizer periodically performs open-vocabulary detection on key frames, while the Segmenter generates and propagates instance masks within short video segments and resets its internal state at segment boundaries to constrain memory and computational costs. We further design Lifecycle-aware ID Association (LIA), which associates cross-segment local trajectories at the output level, maintains globally consistent video identities, and unifies segmented local trajectories into the instance-level segmentation trajectories required by UAV-OVVIS. This design reduces computational overhead, improves open-vocabulary detection coverage in long videos, alleviates missed detections, and stabilizes cross-segment instance identities.

AeroTrack supports modular combinations of Recognizers and Segmenters, where the Recognizer is required to provide open-vocabulary detection and the Segmenter is required to support instance-level segmentation and propagation. In this paper, we use YOLO-World~\cite{cheng2024yoloworld}, Grounding DINO~\cite{liu2023grounding}, and SAM3~\cite{carion2025sam3} as Recognizers, and SAM2~\cite{ravi2024sam2} and SAM3 as Segmenters, instantiating five representative combinations.

Moreover, considering the severe scarcity of instance-level annotations in existing UAV scenarios~\cite{huang2025uavvis}, we construct AeroVIS, a dataset covering 9 UAV object categories and 8,279 instance-level trajectories, to quantitatively evaluate UAV-OVVIS. We further build a UAV-adapted evaluation protocol based on existing VIS benchmark protocols~\cite{wang2023towards}, providing a reproducible quantitative evaluation benchmark for UAV-OVVIS.

Our contributions are summarized as follows:
\begin{itemize}
    \item We introduce UAV-OVVIS, a new task for \textbf{4W} UAV Visual Perception (Figure~\ref{fig:motivation}), which requires discovering targets in UAV videos according to open-vocabulary queries and outputting instance-level segmentation trajectories with globally consistent video identities.
    \item Considering the scarcity of instance-level annotations in UAV scenarios, we reuse existing VFMs to build AeroTrack, a unified training-free framework, and instantiate five feasible variants for enabling UAV-OVVIS.
    \item We construct AeroVIS, a dataset covering 9 UAV object categories and 8,279 trajectories, together with a corresponding evaluation protocol, providing a reproducible quantitative evaluation benchmark for UAV-OVVIS.
    \item Experiments on AeroVIS and three general VIS benchmark datasets show that AeroTrack substantially outperforms existing general video instance segmentation methods in UAV scenarios, demonstrating strong open-vocabulary robustness and generalization.
\end{itemize}

\begin{figure*}[t]
\centering
\includegraphics[width=0.98\textwidth]{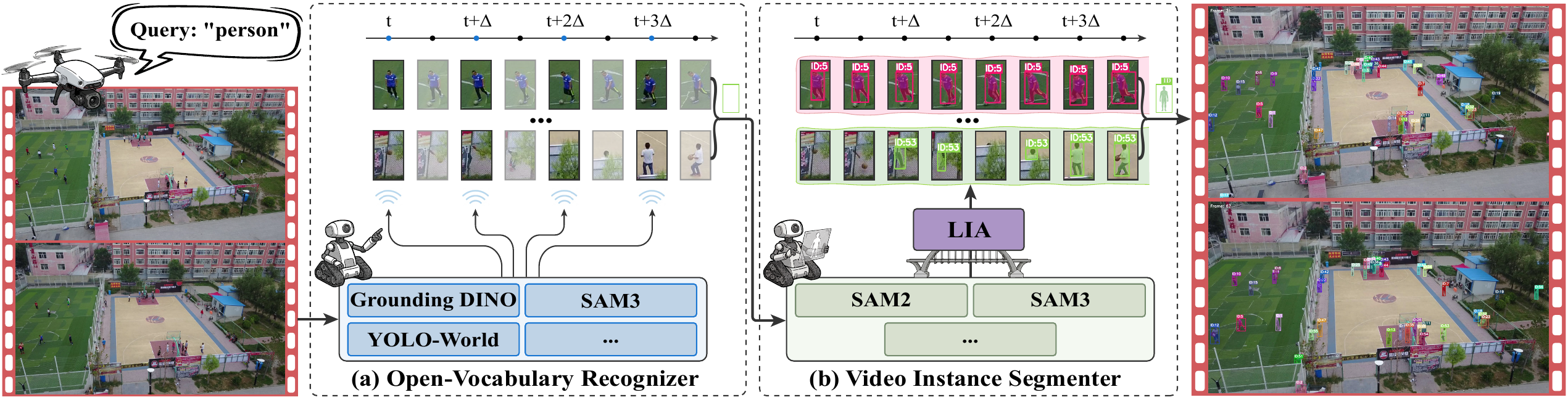}
\caption{Overview of the AeroTrack framework. Given a UAV video and open-vocabulary text queries, (a) the \textbf{Recognizer} uses Grounding DINO, YOLO-World, SAM3, or other models as the open-vocabulary detection module, performs open-vocabulary detection every $\Delta$ frames, and passes positional prompts to the Segmenter. (b) The \textbf{Segmenter} uses SAM2, SAM3, or other promptable video segmenters to perform instance-level segmentation and propagation within each $[k,\bar{k})$ segment according to the positional prompts. It outputs instance masks with local IDs and resets the internal Segmenter memory at each $\Delta$ boundary. \textbf{LIA}, as an output-level association module, further associates local IDs from different $\Delta$ intervals into video-level global IDs, finally producing instance-level segmentation trajectories with globally consistent video identities.}
\label{fig:framework}
\end{figure*}

\section{Related Work}

\noindent\textbf{Open-Vocabulary Recognition.}
Open-vocabulary recognition aims to enable models to discover and localize targets in visual content according to text queries, without being constrained by a predefined category vocabulary. Vision-language contrastive models represented by CLIP align visual and textual representations at the image level through large-scale image-text pretraining, laying the foundation for this capability~\cite{radford2021learning}. Building on this paradigm, the alignment granularity has been extended from images to regions. GLIP formulates object detection as phrase grounding and learns region-language correspondence under a grounded pretraining framework~\cite{li2022glip}. Grounding DINO follows the grounded language-image pretraining paradigm and integrates text-conditioned localization with an end-to-end Transformer detector, improving open-set object localization~\cite{liu2023grounding}. YOLO-World embeds open-vocabulary recognition into the YOLO detection paradigm, preserving text-conditioned detection while maintaining inference efficiency~\cite{cheng2024yoloworld}. Open-vocabulary detectors represented by Grounding DINO and YOLO-World provide the open-vocabulary detection capability required by UAV-OVVIS.

\noindent\textbf{Promptable Segmentation.}
Promptable segmentation aims to generate category-agnostic image instance masks from positional prompts such as points, boxes, or masks. SAM pioneered this paradigm and achieved such capability through large-scale pretraining~\cite{kirillov2023segment}. SAM2 extends promptable segmentation to the video domain, retaining image segmentation capability while enabling cross-frame mask propagation through streaming memory~\cite{ravi2024sam2}. SAM3 further introduces textual or visual concept prompts, jointly performing detection, segmentation, and tracking of matched instances in images and videos~\cite{carion2025sam3}. SEEM realizes open-vocabulary interactive segmentation with compositional multimodal prompts~\cite{zou2023seem}. SAM-I2V upgrades pretrained SAM to promptable video segmentation with relatively low training cost~\cite{mei2025sami2v}. Promptable segmentation models represented by SAM2 and SAM3 provide instance segmentation and temporal propagation capabilities for UAV-OVVIS.

\noindent\textbf{UAV Video Perception.}
UAV video perception has mainly focused on box-level localization and trajectory association, while gradually extending to pixel-level segmentation and instance-level trajectory modeling. For box-level tasks, VisDrone covers object detection, single-object tracking, and multi-object tracking in UAV images and videos, providing a unified evaluation protocol for small objects and multi-category targets from low-altitude viewpoints~\cite{zhu2018vision}. UAVDT systematically evaluates vehicle detection and multi-object tracking in complex traffic scenarios, highlighting the impact of camera motion, scale variation, and occlusion on trajectory association~\cite{du2018uav}. In addition, WebUAV-3M supports large-scale UAV single-object tracking evaluation with million-frame scale data~\cite{zhang2022webuav3m}. For pixel-level segmentation, iSAID provides large-scale aerial instance segmentation and highlights the challenges of small objects and dense occlusions~\cite{waqas2019isaid}. UAVid constructs an oblique-view UAV video semantic segmentation benchmark~\cite{lyu2018uavid}. VDD integrates multi-source UAV imagery to expand scene coverage for semantic segmentation~\cite{cai2023vdd}. For instance-level understanding, HT-VIS jointly models cross-frame mask propagation and instance identity through hierarchical offset compensation and temporal memory updates, introducing closed-set VIS into UAV videos~\cite{huang2025uavvis}.

\noindent\textbf{Open-Vocabulary Video Instance Segmentation.}
Open-Vocabulary Video Instance Segmentation aims to discover specified targets according to open-vocabulary text queries and output video instance segmentation trajectories. OV2Seg first introduced the Open-Vocabulary Video Instance Segmentation (OV-VIS) task, established the LV-VIS benchmark, and achieved end-to-end open-vocabulary instance trajectory prediction through a Memory-Induced Transformer~\cite{wang2023towards}. DEVA decouples image-level segmentation from temporal propagation, providing a reusable solution for modular OV-VIS~\cite{cheng2023tracking}. OVFormer mitigates the domain gap between instance queries and CLIP representations through unified embedding alignment, and improves OV-VIS performance by exploiting video temporal consistency~\cite{fang2024ovformer}. CLIP-VIS adapts frozen CLIP to OV-VIS, combining class-agnostic mask generation with temporal matching to improve open-vocabulary generalization~\cite{zhu2024clipvis}. OpenVIS further systematizes OV-VIS evaluation protocols and proposes the InstFormer framework for open-category video instance segmentation~\cite{guo2025openvis}. Unified foundation models such as GLEE and SAM3 support detection, segmentation, and tracking in images and videos through general object representations and concept prompts, respectively, providing reusable unified model capabilities for OV-VIS~\cite{wu2023glee,carion2025sam3}. However, existing OV-VIS benchmarks and methods mainly target natural-view videos and struggle to stably generate high-quality instance-level segmentation trajectories in complex UAV scenarios.

\section{Problem Definition}
\label{sec:problem-def}
UAV-OVVIS aims to discover specified targets in UAV videos according to open-vocabulary text queries and output instance-level segmentation trajectories with globally consistent video identities. Existing UAV video benchmarks are mostly based on box-level detection or tracking annotations, while instance-level mask annotations remain scarce, making large-scale supervised training difficult~\cite{zhu2018vision,du2018uav,huang2025uavvis}. Meanwhile, existing visual foundation models (VFMs), typically pretrained on large-scale natural images and videos, already possess open-vocabulary detection, instance-level segmentation, and propagation capabilities~\cite{liu2023grounding,ravi2024sam2,carion2025sam3}. Although UAV viewpoints introduce domain gaps such as small objects, top-down perspectives, and camera motion, the general visual representations learned by these models still have the potential to transfer across scenarios~\cite{chen2024uavdb,blei2025cloudtrack}. Therefore, this paper focuses on reusing existing VFMs in a training-free manner to enable stable and high-quality UAV-OVVIS without UAV-domain annotations or retraining. This problem mainly faces the following challenges.
\begin{itemize}
    \item \textbf{Capability Coordination among VFMs.} UAV-OVVIS simultaneously requires open-vocabulary detection, instance-level segmentation and propagation, and global identity maintenance. How to decouple UAV-OVVIS into cooperative foundation-vision tasks under a training-free setting and reuse different VFMs to build a unified inference framework is the primary issue in enabling this task.

    \item \textbf{Temporal Target Coverage and Identity Unification in UAV Videos.} In UAV videos, the platform and targets often move simultaneously, while objects are small, densely distributed, and frequently occluded. UAV-OVVIS must not only continuously discover newly entering or reappearing small objects in long videos, but also distinguish new objects from historical ones and maintain globally consistent video identities.

    \item \textbf{Resource Pressure in Long Sequences.} When performing open-vocabulary detection, instance-level segmentation, and propagation, existing VFMs usually incur higher memory and computational costs than conventional image-level or closed-set tasks. For long UAV video sequences with a large number of parallel small objects, this pressure is further amplified, limiting the processable video length and target scale.
\end{itemize}

\section{Methodology}
\label{sec:method}

To address the above challenges, we follow the training-free composition paradigm of grounding-guided segmentation and extend it into a staged inference pipeline, where text-conditioned open-vocabulary detection first generates positional prompts and then drives video segmentation and propagation~\cite{ren2024groundedsam,cheng2023tracking}. However, directly adopting this pipeline cannot simultaneously satisfy the requirements of UAV-OVVIS for temporal target coverage, long-sequence resource efficiency, and global identity maintenance. To this end, AeroTrack adopts the design principles of periodic target refreshing, short-segment temporal propagation, and global instance identity unification. It decouples UAV-OVVIS into two components: an Open-Vocabulary Recognizer (Recognizer) and a Video Instance Segmenter (Segmenter), which respectively perform key-frame open-vocabulary detection and within-segment mask propagation. We further design Lifecycle-aware ID Association (LIA) to maintain globally consistent video identities across segments. Periodic refreshing and state resetting at segment boundaries can control long-sequence costs, but they also interrupt the cross-segment memory of the Segmenter, causing the same physical target to be assigned new local IDs in different segments. Therefore, LIA re-associates local trajectories into video-level global instance trajectories at the output level according to lifecycle states and geometric consistency.

\subsection{AeroTrack Framework}

As shown in Figure~\ref{fig:framework}, AeroTrack takes a UAV video $V=\{I_t\}_{t=0}^{T-1}$ and an open-vocabulary text query set $\mathcal{C}$ as input, and outputs instance-level segmentation trajectories with globally consistent video identities. The system defines a key-frame set $\mathcal{K}=\{0,\Delta,2\Delta,\ldots\}\cap[0,T)$ with a fixed refresh interval $\Delta$, and accordingly divides the video into local temporal segments. For each key frame $k\in\mathcal{K}$, let $\bar{k}=\min(k+\Delta,T)$. The Recognizer first performs text-conditioned open-vocabulary detection on $I_k$ according to $\mathcal{C}$, generating a positional prompt set $\mathcal{P}_k$. The Segmenter then takes $\mathcal{P}_k$ as initialization prompts, generates and propagates instance masks within the segment $[k,\bar{k})$, and produces local trajectories with local IDs $\ell$. Its internal state is reset at segment boundaries to constrain the effective memory length. Finally, LIA receives the Segmenter outputs and establishes an output-level mapping from local IDs $\ell$ to video-level global IDs $g$, thereby associating segment-level local trajectories into video instance trajectories with globally consistent identities.

\subsection{Open-Vocabulary Recognizer}

As shown in Figure~\ref{fig:framework}(a), the Recognizer is implemented by VFMs that support open-vocabulary detection and output positional prompts. We select YOLO-World, Grounding DINO, and SAM3 as representative Recognizers. Given an open-vocabulary text query set $\mathcal{C}$, the Recognizer performs text-conditioned open-vocabulary detection every $\Delta$ frames on the key frame $I_k$ corresponding to $k\in\mathcal{K}$. To control redundant prompts and the scale of parallel targets, candidate results are processed by confidence filtering, spatial deduplication, and target-budget truncation to form the positional prompt set
\begin{equation}
\label{eq:recognizer}
\mathcal{P}_k=\operatorname{Rec}(I_k,\mathcal{C})
=\{(p_j,c_j,s_j)\}_{j=1}^{N_k},\quad k\in\mathcal{K},
\end{equation}
where $p_j$ denotes the positional prompt of a target matched to the text query, $c_j$ denotes the matched text category, and $s_j$ denotes the matching confidence. $\mathcal{P}_k$ is then passed to the Segmenter to initialize instance-level segmentation and propagation in the corresponding temporal segment.

\subsection{Video Instance Segmenter}

As shown in Figure~\ref{fig:framework}(b), the Segmenter is implemented by VFMs that support positional-prompt input, instance-level segmentation, and propagation. We select SAM2 and SAM3 as representative Segmenters. For a segment starting from $k$, let $\bar{k}=\min(k+\Delta,T)$ and $V_{[k,\bar{k})}=\{I_t\}_{t=k}^{\bar{k}-1}$. According to the positional prompt set $\mathcal{P}_k$ produced by the Recognizer, the Segmenter generates and propagates instance masks within $V_{[k,\bar{k})}$, and outputs local trajectories within the segment:
\begin{equation}
\mathcal{M}_k = \mathrm{Seg}\!\left(\mathcal{P}_k,\,V_{[k,\,\bar{k})}\right)
= \left\{\left(\{m_{\ell,t}\}_{t=k}^{\bar{k}-1},\,\mathrm{id}_\ell^{\mathrm{local}}\right)\right\}_\ell,
\end{equation}
where $m_{\ell,t}$ denotes the mask of the $\ell$-th local instance in frame $t$, and $\mathrm{id}_\ell^{\mathrm{local}}$ denotes the local ID assigned by the Segmenter within the current segment. Since this local ID is valid only within the current segment, it cannot be directly used as a globally consistent video identity.

At each $\Delta$ boundary, AeroTrack resets the internal tracking state of the Segmenter so that it only maintains target memory within the current $[k,\bar{k})$. As a result, the effective memory length of the Segmenter is reduced from the entire video to at most $\Delta$, alleviating the temporal memory pressure caused by long sequences and dense targets. The local IDs and instance masks output by the Segmenter are then jointly passed to LIA.

\subsection{Lifecycle-aware ID Association}
\label{sec:lia}

LIA is a lightweight module that bridges the identity fragmentation introduced by segmented propagation. It introduces no trainable modules and does not modify the internal propagation state of the Segmenter. Instead, it maps within-segment local IDs to video-level global IDs at the output level according to lifecycle states and geometric consistency. Since the Segmenter is reset at every $\Delta$ boundary, the same physical target may be assigned new local IDs in different segments. LIA therefore restores cross-segment identity consistency by maintaining short-term lifecycle states. It should be noted that LIA is neither an appearance re-identification module nor a long-term tracking module. Its role is to repair local ID fragmentation caused by segment restarts, short-term occlusions, or propagation interruptions between periodic detection and segmented propagation. When a target is not re-detected for a long period, its historical identity expires after $T_{\mathrm{lost}}$. The complete procedure is summarized in Algorithm~\ref{alg:lia}.

\begin{algorithm}[t]
\caption{Lifecycle-aware ID Association (LIA)}
\label{alg:lia}
\small
\begin{algorithmic}[1]
\Require Segment index $s$; per-frame Segmenter outputs $\{\mathcal{O}_t\}_{t\in s}$ with $\mathcal{O}_t=\{(m,\ell)\}$; lifecycle state $\Omega=(\mathcal{B}_{\mathrm{act}},\mathcal{B}_{\mathrm{lost}},\Phi,\mathrm{gid}_{\mathrm{next}})$
\Ensure Global IDs $\{g\}$ and updated lifecycle state $\Omega$
\For{each frame $t$ in segment $s$}
    \State Extract valid boxes $\mathcal{D}_t=\{(\ell,b_\ell)\}$ from $\mathcal{O}_t$ by \textsc{ValidBBox}
    \State $\mathcal{U} \gets \{\ell \mid (\ell,b_\ell)\in\mathcal{D}_t \land (s,\ell) \notin \Phi\}$
    \If{$\mathcal{U} \neq \emptyset$}
        \State Build candidates $(\ell,g)$ from $\mathcal{U}$ and $\mathcal{B}_{\mathrm{act}}\cup\mathcal{B}_{\mathrm{lost}}$ that pass \textsc{GeomGate}
        \State Compute $S$ by Eq.~\eqref{eq:lia-score}, then derive threshold score $\widetilde{S}$ and ranking score $\hat{S}$
        \State Keep candidates with $\widetilde{S}\geq\tau_{\mathrm{life}}$
        \State Greedily assign disjoint pairs by descending $\hat{S}$
        \State Set $\Phi(s,\ell)\gets g$ for each matched pair $(\ell,g)$
        \State Allocate a new global ID for each unmatched $\ell\in\mathcal{U}$ and update $\Phi$
    \EndIf
    \For{each $(\ell,b_\ell)\in\mathcal{D}_t$ with $(s,\ell)\in\Phi$}
        \State \textsc{Observe}$(\Phi(s,\ell),b_\ell)$
    \EndFor
\EndFor
\State Update $\mathcal{B}_{\mathrm{act}}$ and $\mathcal{B}_{\mathrm{lost}}$ at segment end
\State Remove expired tracks from $\mathcal{B}_{\mathrm{lost}}$
\end{algorithmic}
\end{algorithm}

LIA maintains a lifecycle bank $\mathcal{B}=\mathcal{B}_{\mathrm{act}}\cup\mathcal{B}_{\mathrm{lost}}$, where $\mathcal{B}_{\mathrm{act}}$ records currently visible global trajectories and $\mathcal{B}_{\mathrm{lost}}$ records historical trajectories that are temporarily unobserved but still recoverable. Each trajectory stores a global ID, the most recent valid bounding box, and the number of lost segments. For a local trajectory $\ell$ in the current segment, LIA converts its valid mask into a bounding box $b_\ell$ and matches it with the historical bounding box $b_g$ of a candidate global trajectory $g$. Hard gating first filters out candidates with unreasonable scale changes or displacement, requiring the area ratio $A(b_\ell)/A(b_g)$ to fall within $[\rho_{\min},\rho_{\max}]$ and the center distance to be no larger than $\gamma\sqrt{\max(A(b_\ell),A(b_g))}$. After gating, the geometric matching score is computed as
\begin{equation}
\label{eq:lia-score}
\begin{aligned}
S(b_\ell,b_g)=
&\,0.45\,\operatorname{IoU}(b_\ell,b_g)
+0.30\,C(b_\ell,b_g)\\
&+0.25\,A_{\mathrm{sim}}(b_\ell,b_g),
\end{aligned}
\end{equation}
where the three terms measure spatial overlap, center similarity, and scale consistency, respectively. Let $A_\ell=A(b_\ell)$ and $A_g=A(b_g)$, and let $d(b_\ell,b_g)$ denote the center distance between the two boxes. The center similarity and area similarity are defined as
\begin{equation}
\label{eq:lia-similarity}
\begin{aligned}
C(b_\ell,b_g)&=\max\!\left(0,1-\frac{d(b_\ell,b_g)}{\gamma\sqrt{\max(A_\ell,A_g)}}\right),\\
A_{\mathrm{sim}}(b_\ell,b_g)&=\frac{\min(A_\ell,A_g)}{\max(A_\ell,A_g)}.
\end{aligned}
\end{equation}
Based on $S$, LIA computes a lifecycle score $\widetilde{S}$, where \textit{lost} trajectories are decayed according to the number of lost segments, while \textit{active} trajectories keep $\widetilde{S}=S$. Candidate matches are jointly filtered by geometric gating and $\widetilde{S}$, and a ranking score $\hat{S}$ further gives a slight priority to \textit{active} trajectories. LIA then performs one-to-one greedy matching in descending order of $\hat{S}$, where each local ID and each global ID can be matched at most once in the same round. Matched local trajectories inherit the corresponding global IDs, while unmatched trajectories are assigned new global IDs. Valid observations within a segment continuously update the lifecycle bank, supporting identity continuation under local ID jitter and providing up-to-date geometric states for short-term lost-track recovery in subsequent segments.

\section{Experiments}
\label{sec:experiments}

\begin{table*}[!t]
\centering
\scriptsize
\setlength{\tabcolsep}{2.05pt}
\renewcommand{\arraystretch}{1.05}
\resizebox{\textwidth}{!}{
\begin{tabular}{@{}l
c@{\hspace{2.4pt}}c@{\hspace{2.4pt}}c@{\hspace{2.4pt}}c@{\hspace{2.4pt}}c@{\hspace{2.4pt}}c@{\hspace{2.4pt}}c@{\hspace{2.4pt}}c@{\hspace{2.4pt}}c@{\hspace{7pt}}
c@{\hspace{2.4pt}}c@{\hspace{2.4pt}}c@{\hspace{2.4pt}}c@{\hspace{2.4pt}}c@{}}
\toprule
\multirow{2}{*}{\textbf{Model}} & \multicolumn{9}{c}{\textit{Per-category} \textbf{HOTA}} & \multicolumn{5}{c}{\textit{Overall (GT-track weighted)}} \\
\cmidrule(lr){2-10}\cmidrule(lr){11-15}
 & \textbf{Person} & \textbf{Car} & \textbf{Truck} & \textbf{Bus} & \textbf{Bicycle} & \textbf{Moto.} & \textbf{Tricycle} & \textbf{Boat} & \textbf{Vehicle} & mAP & DetA & AssA & Mask J & \textbf{HOTA} \\
\cmidrule(lr){2-2}\cmidrule(lr){3-3}\cmidrule(lr){4-4}\cmidrule(lr){5-5}\cmidrule(lr){6-6}\cmidrule(lr){7-7}\cmidrule(lr){8-8}\cmidrule(lr){9-9}\cmidrule(lr){10-10}\cmidrule(lr){11-11}\cmidrule(lr){12-12}\cmidrule(lr){13-13}\cmidrule(lr){14-14}\cmidrule(lr){15-15}
DEVA~\cite{cheng2023tracking} & 29.17 & 52.90 & 23.88 & 30.12 & 16.31 & 27.97 & 19.22 & 77.16 & 17.83 & 10.99 & 24.64 & 53.74 & 39.97 & 34.98 \\
GLEE~\cite{wu2023glee} & 17.14 & 35.35 & 9.08 & 10.06 & 5.79 & 9.89 & 9.54 & 17.28 & 31.14 & 3.52 & 16.75 & 38.36 & 47.12 & 24.71 \\
OV2Seg~\cite{wang2023towards} & 2.53 & 15.20 & 0.00 & 8.23 & 0.00 & 5.31 & 1.31 & 56.48 & 0.00 & 0.40 & 3.76 & 27.26 & 12.93 & 6.97 \\
CLIP-VIS~\cite{zhu2024clipvis} & 3.03 & 13.49 & 3.44 & 6.11 & 1.81 & 3.00 & 4.14 & 12.72 & 18.54 & 0.45 & 4.95 & 20.54 & 16.24 & 9.55 \\
OVFormer~\cite{fang2024ovformer} & 2.44 & 7.05 & 3.17 & 6.87 & 0.70 & 2.40 & 1.97 & 31.16 & 12.44 & 0.42 & 1.57 & 25.79 & 8.18 & 5.94 \\
SAM3~\cite{carion2025sam3} & 18.72 & 19.03 & 43.04 & 65.97 & 25.57 & 23.55 & 38.53 & 81.37 & 41.93 & 5.43 & 12.29 & 56.83 & \textbf{67.18} & 25.61 \\[-2pt]
\multicolumn{15}{@{}c@{}}{{\fontsize{5}{5}\selectfont\itshape UAV-OVVIS (Ours)}} \\[-1pt]
YOLO-World + SAM2 & 39.33 & 59.51 & 20.78 & 35.15 & 23.28 & 34.91 & 25.39 & 82.18 & 53.61 & 22.05 & 35.71 & 64.97 & 52.26 & 47.55 \\
Grounding DINO + SAM2 & 38.92 & 62.53 & 23.05 & 47.39 & 21.95 & 34.19 & 25.75 & 84.02 & 52.89 & 21.48 & 37.13 & 65.00 & 54.74 & 48.42 \\
SAM3 + SAM3 & 44.79 & 63.35 & \textbf{50.44} & \textbf{67.58} & \textbf{30.21} & 39.37 & \textbf{51.78} & \textbf{84.32} & 63.49 & 22.19 & 44.48 & \textbf{67.61} & 61.71 & 54.43 \\
YOLO-World + SAM3 & 45.15 & \textbf{66.34} & 47.72 & 66.60 & 28.83 & \textbf{39.57} & 47.45 & 84.27 & \textbf{66.03} & 23.10 & \textbf{47.02} & 67.08 & 57.66 & \textbf{55.82} \\
Grounding DINO + SAM3 & \textbf{45.17} & 65.69 & 48.05 & 62.70 & 30.02 & 37.13 & 48.17 & 83.73 & 65.73 & \textbf{24.46} & 46.44 & 66.89 & 58.65 & 55.39 \\
\bottomrule
\end{tabular}
}
\caption{Comparison results on the AeroVIS dataset. The upper block reports existing general OV-VIS methods transferred to UAV scenarios, while the lower block reports our five AeroTrack variants. Per-category columns report HOTA. Overall metrics aggregate mAP, DetA, AssA, Mask J, and HOTA by weighting each category according to the number of GT tracks. \textbf{Bold} indicates the best result in each column.}
\label{tab:external-methods}
\end{table*}

\begin{figure}[t]
\centering
\includegraphics[width=\columnwidth]{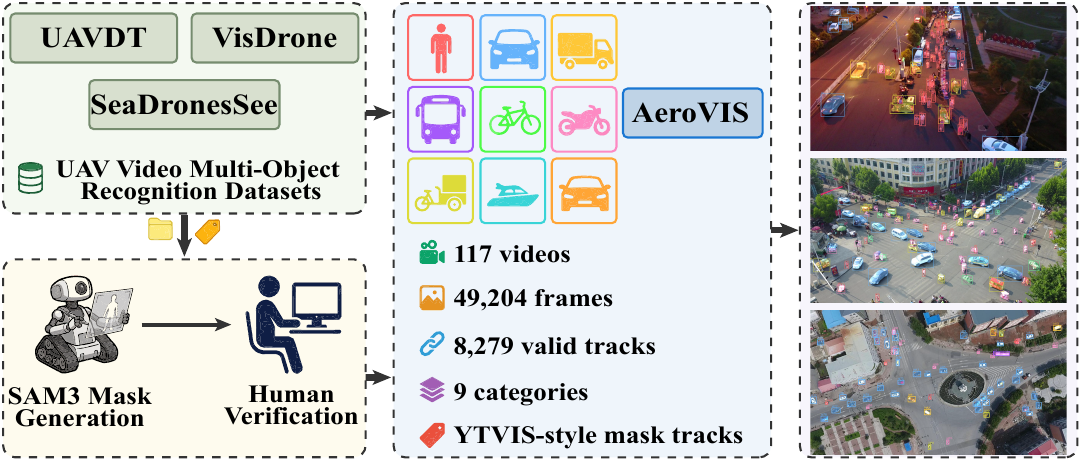}
\caption{AeroVIS dataset construction pipeline. We use box-level annotations from existing UAV multi-object tracking data as positional and identity priors, and combine SAM3 mask generation, temporal organization, and manual reinspection to construct YTVIS-style instance-level segmentation trajectories.}
\label{fig:aerovis-construction}
\end{figure}

\subsection{Dataset}
\label{sec:exp-setup}
\noindent Existing UAV scenarios lack evaluation benchmarks for open-vocabulary video instance segmentation. Therefore, we construct AeroVIS as the main benchmark for UAV-OVVIS, and further evaluate AeroTrack on three general VIS datasets, YouTube-VIS 2019, YouTube-VIS 2021~\cite{yang2019video}, and LV-VIS~\cite{wang2023towards}. All external datasets are used only for evaluation and are not involved in training or fine-tuning.

\noindent\textbf{AeroVIS.} As shown in Figure~\ref{fig:aerovis-construction}, we construct AeroVIS based on UAV multi-object tracking datasets such as VisDrone~\cite{zhu2018vision}, UAVDT~\cite{du2018uav}, and SeaDronesSee~\cite{varga2022seadronessee}. AeroVIS uses frame-level manual box annotations from the source datasets as positional and identity priors to drive SAM3 mask candidate generation, and obtains YTVIS-style instance-level segmentation trajectories through temporal organization, quality filtering, and manual reinspection. AeroVIS contains 117 videos, 49,204 frames, 9 UAV object categories, and 8,279 valid trajectories.

\noindent\textbf{YouTube-VIS 2019.} YouTube-VIS 2019~\cite{yang2019video} is the first large-scale VIS benchmark, containing 2,883 natural videos, 40 categories, and approximately 131k instance masks. We use its validation set for evaluation, which contains 302 videos, 8,289 frames, and 511 instance trajectories.

\noindent\textbf{YouTube-VIS 2021.} YouTube-VIS 2021 follows the 40-class closed-set category space of YouTube-VIS and provides a subsequent general VIS evaluation protocol. We use its validation set for evaluation, which contains 421 videos, 13,195 frames, and 898 instance trajectories.

\noindent\textbf{LV-VIS.} LV-VIS~\cite{wang2023towards} is a large-vocabulary video instance segmentation benchmark for OV-VIS, covering 1,196 categories and thus substantially larger category scale than conventional VIS datasets. We use its validation set for evaluation, which contains 837 videos, 19,139 frames, and 3,719 instance trajectories.

\subsection{Implementation Details}
\noindent\textbf{Codebase.} We will release the complete AeroTrack project. To the best of our knowledge, this is the first open-source framework in the UAV domain that systematically supports open-vocabulary video instance segmentation. The project is built around the Recognizer, Segmenter, and LIA modules, provides five training-free variants, and includes the AeroVIS dataset, unified inference scripts, and complete evaluation tools to support the reproduction, comparison, and future research of UAV-OVVIS.

\noindent\textbf{Setup.} All reproduction and inference experiments are conducted on a single NVIDIA RTX 3090 (24\,GB) GPU. We evaluate five training-free variants without any training or fine-tuning. Since AeroTrack is designed for diverse UAV scenarios, we adopt unified inference settings, configuring only the single thresholds of the Recognizer and Segmenter at the category level while keeping all other hyperparameters unchanged. The open-vocabulary detection refresh interval is set to $\Delta=25$ frames, which jointly considers inference efficiency, segment propagation stability, and practical detection coverage. The corresponding sensitivity analysis is provided in the supplementary material. Text prompts use the corresponding category names. SAM3-related combinations use publicly available pretrained weights, SAM2-related combinations use SAM2.1 Hiera-L, YOLO-World-related combinations use YOLO-World-L, and Grounding DINO-related combinations use Grounding DINO SwinT.

\noindent\textbf{Evaluation.} In AeroVIS experiments, we use HOTA as the primary metric, and report its DetA and AssA components~\cite{luiten2021hota}, mAP~\cite{yang2019video,wang2023towards}, and Mask J~\cite{perazzi2016benchmark} as supplementary metrics. HOTA jointly measures detection coverage and temporal identity association quality, with DetA and AssA corresponding to detection accuracy and identity stability, respectively. mAP follows the official YouTube-VIS/LV-VIS protocol and computes AP for instance-level segmentation trajectories. Mask J denotes the average mask Jaccard/IoU of matched trajectories.

\noindent\textbf{Compared Methods.}
Considering that public methods specifically designed for video instance segmentation in UAV scenarios remain lacking, we select existing general-scene OV-VIS methods for comparison. In AeroVIS experiments, we reproduce DEVA~\cite{cheng2023tracking}, GLEE~\cite{wu2023glee}, OV2Seg~\cite{wang2023towards}, CLIP-VIS~\cite{zhu2024clipvis}, OVFormer~\cite{fang2024ovformer}, and SAM3~\cite{carion2025sam3}, and compare them with the five AeroTrack variants. All reproduced methods are evaluated under a unified protocol using publicly available checkpoints and recommended configurations.

Beyond AeroVIS, to evaluate the open-vocabulary generalization of AeroTrack, we further conduct experiments on three standard benchmarks, YouTube-VIS 2019, YouTube-VIS 2021, and LV-VIS. Since relatively complete public results are available on these benchmarks, we include representative OV-VIS methods such as DEVA~\cite{cheng2023tracking}, OV2Seg~\cite{wang2023towards}, GLEE~\cite{wu2023glee}, CLIP-VIS~\cite{zhu2024clipvis}, OVFormer~\cite{fang2024ovformer}, BriVIS~\cite{cheng2024brivis}, Troy-VIS~\cite{yan2024troyvis}, OpenVIS~\cite{guo2025openvis}, and SAM3~\cite{carion2025sam3}, and directly quote the metrics reported in their papers or public reports for comparison.

\begin{figure*}[t]
\centering
\includegraphics[width=0.98\textwidth]{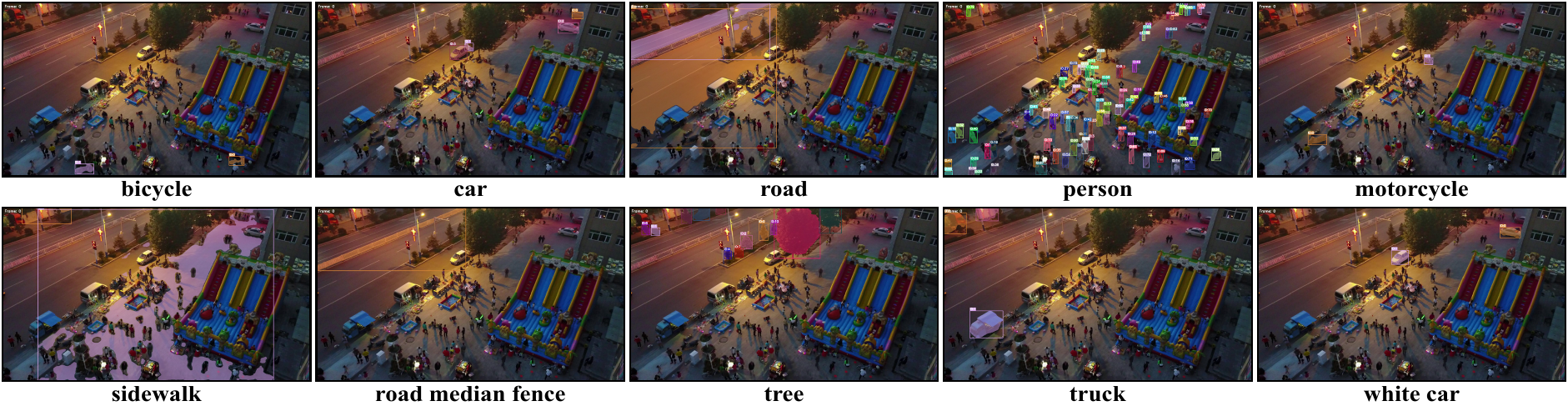}
\caption{Open-vocabulary qualitative results on the same UAV video. Each subfigure corresponds to a set of open-vocabulary text queries, and the text below each subfigure shows the actual input query.}
\label{fig:qualitative-open-vocabulary}
\end{figure*}

\begin{table}[t]
\centering
\scriptsize
\setlength{\tabcolsep}{2.5pt}
\renewcommand{\arraystretch}{0.92}
\resizebox{\columnwidth}{!}{
\begin{tabular}{@{}lccc@{}}
\toprule
\multirow{2}{*}{\textbf{Model}} & \multicolumn{3}{c}{\textit{mAP}} \\
\cmidrule(lr){2-4}
 & \textbf{YTVIS19} & \textbf{YTVIS21} & \textbf{LV-VIS} \\
\midrule
DEVA~\cite{cheng2023tracking} & 40.8 & -- & -- \\
OV2Seg~\cite{wang2023towards} & 37.6 & 33.9 & 21.1 \\
GLEE~\cite{wu2023glee} & \textbf{63.6} & -- & 30.3 \\
CLIP-VIS~\cite{zhu2024clipvis} & 42.3 & 39.5 & 32.1 \\
OVFormer~\cite{fang2024ovformer} & 44.3 & 37.6 & 24.7 \\
BriVIS~\cite{cheng2024brivis} & 45.3 & 39.5 & 27.7 \\
Troy-VIS~\cite{yan2024troyvis} & -- & -- & 20.9 \\
OpenVIS~\cite{guo2025openvis} & -- & -- & 14.1 \\
SAM3~\cite{carion2025sam3} & -- & \textbf{57.4} & 36.3$^{*}$ \\[-2pt]
\multicolumn{4}{c}{{\fontsize{5}{5}\selectfont\itshape UAV-OVVIS (Ours)}} \\[-1pt]
YOLO-World + SAM2 & 36.22 & 32.60 & 29.88 \\
Grounding DINO + SAM2 & 39.01 & 30.91 & 21.21 \\
SAM3 + SAM3 & 56.64 & 56.29 & \textbf{36.47} \\
YOLO-World + SAM3 & 47.70 & 49.31 & 35.63 \\
Grounding DINO + SAM3 & 47.74 & 48.11 & 34.97 \\
\bottomrule
\end{tabular}
}
\caption{mAP comparison on three general VIS benchmarks. Results of public methods are listed according to their original papers or public reports, while AeroTrack results are evaluated on the complete validation set of each dataset following the official protocol. \textbf{Bold} indicates the best result in each column. $^{*}$ indicates that the result is from the test split, while all other results are from the val split.}
\label{tab:general-benchmarks}
\vspace{-0.08in}
\end{table}

\subsection{Comparison on AeroVIS}

As shown in Table~\ref{tab:external-methods}, we conduct comparison experiments on AeroVIS under a unified protocol. Since no dedicated method currently exists for UAV-OVVIS, we reproduce general-scene methods capable of OV-VIS to evaluate their practical transfer performance in UAV scenarios, and compare them with the five AeroTrack variants. Overall, existing methods show limited performance after being transferred to UAV scenarios. A notable phenomenon is that the DetA of existing methods is generally lower than their AssA, indicating that detection coverage for text-specified targets is a more prominent bottleneck than short-term identity association for already detected targets. For dense small objects, frequent entry and exit from the field of view, and occlusion-induced reappearance in UAV videos, general methods struggle to promptly discover newly entering targets and stably recover trajectories after targets are lost or reappear, which limits their performance on AeroVIS. It is also worth noting that the five external methods other than Native SAM3 are relatively lightweight and do not reach the memory limit during inference. In contrast, Native SAM3 is constrained by the number of detections and the video length, which we further analyze in \S\ref{sec:memory-exp}. Therefore, when reproducing Native SAM3, we set the number of detections to the maximum target budget $N$ supported under the memory constraint for each category. Native SAM3 performs close to AeroTrack on categories with fewer targets, such as boat and bus, but its performance drops significantly on categories with many same-frame instances, such as person and car. This suggests that although SAM3 has strong concept-prompting and instance segmentation capabilities, it remains limited by target budget, memory growth, and re-detection capability in long UAV videos with dense targets, which is precisely the motivation of this work.

By contrast, all five AeroTrack variants substantially outperform both external methods and Native SAM3, demonstrating that periodic open-vocabulary detection, short-segment propagation, and LIA can effectively adapt to UAV-OVVIS. Since the Recognizer and Segmenter are independent in AeroTrack, and all variants use unified inference settings except for category-level control thresholds, different combinations are fairly comparable. Comparing the five variants shows that using SAM3 as the Segmenter generally outperforms SAM2, reflecting stronger instance-level segmentation and propagation capabilities. YOLO-World and Grounding DINO as Recognizers slightly outperform SAM3 concept-prompt detection, indicating that different Recognizers lead to different performance characteristics. These results validate the stability and replaceability of AeroTrack's modular design.

\subsection{Experiments on General Datasets} 

To further validate the open-vocabulary generalization of AeroTrack, we conduct supplementary evaluation on three general VIS benchmarks, YouTube-VIS 2019/2021 and LV-VIS. In addition to the methods introduced in the AeroVIS comparison, we further include general OV-VIS methods such as BriVIS~\cite{cheng2024brivis}, Troy-VIS~\cite{yan2024troyvis}, and OpenVIS~\cite{guo2025openvis}, and quote their publicly reported metrics. All five AeroTrack variants are re-evaluated on the complete validation sets following the official mAP protocol.

As shown in Table~\ref{tab:general-benchmarks}, AeroTrack maintains stable performance on the three general VIS benchmarks, indicating that the periodic refreshing, short-segment propagation, and LIA designed for long UAV videos do not weaken the general open-vocabulary capability of the integrated VFMs. In particular, on benchmarks covered by the public SAM3 report, SAM3 + SAM3 performs close to Native SAM3, showing that AeroTrack can well inherit SAM3's concept prompting, instance segmentation, and video propagation capabilities. Further comparison among different combinations shows that using SAM3 as the Segmenter generally outperforms SAM2, while using YOLO-World or Grounding DINO as an external Recognizer performs slightly lower than SAM3 concept-prompt detection. This is consistent with the characteristics of general VIS benchmarks, whose sequences are shorter, targets are relatively sparse, and data distribution is closer to SAM3 pretraining. Overall, while enhancing adaptation to UAV scenarios, AeroTrack still preserves the open-vocabulary generalization capability of the integrated VFMs.

\subsection{Open-Vocabulary Qualitative Results}
Figure~\ref{fig:qualitative-open-vocabulary} shows open-vocabulary qualitative results of AeroTrack in a complex nighttime UAV scene. This experiment adopts the YOLO-World + SAM3 variant and keeps the video frames and inference configuration fixed, changing only the input text queries to examine open-vocabulary response capability under practical conditions. The results show that AeroTrack can distinguish semantically close or visually similar queries, such as \textit{car} versus \textit{white car}, and \textit{motorcycle} versus \textit{bicycle}. For non-standard or region-specific targets in UAV scenarios, such as \textit{sidewalk}, \textit{road median fence}, \textit{tree}, and \textit{road}, AeroTrack can also generate corresponding instance masks and maintain identity distinctions. These results demonstrate that AeroTrack supports fine-grained open-vocabulary instance understanding in complex UAV scenarios.

\begin{table}[t]
\centering
\scriptsize
\setlength{\tabcolsep}{2.5pt}
\renewcommand{\arraystretch}{0.88}
\resizebox{\columnwidth}{!}{
\begin{tabular}{lccc}
\toprule
\multirow{2}{*}{\textbf{Model}} & \multicolumn{2}{c}{\textbf{LIA}} & \multirow{2}{*}{\textbf{Gain}} \\[-2pt]
\cmidrule(lr){2-3}
 & \textbf{w/o} & \textbf{w/} & \\[-2pt]
\midrule
YOLO-World + SAM2 & 20.24 & 47.55 & +27.31 \\
Grounding DINO + SAM2 & 20.62 & 48.42 & +27.80 \\
SAM3 + SAM3 & 22.72 & 54.43 & +31.71 \\
YOLO-World + SAM3 & 23.29 & 55.82 & +32.53 \\
Grounding DINO + SAM3 & 23.22 & 55.39 & +32.17 \\
\bottomrule
\end{tabular}
}
\caption{LIA ablation on AeroVIS, reporting only Overall HOTA. Gain denotes the improvement in Overall HOTA when enabling LIA compared with disabling LIA.}
\label{tab:lia-ablation}
\vspace{-0.08in}
\end{table}

\subsection{Ablation Study}

LIA is the key component for maintaining globally consistent video identities in AeroTrack. To verify its necessity, we ablate the five AeroTrack variants in Table~\ref{tab:lia-ablation} by disabling the lifecycle association module. With w/o LIA, the Recognizer still refreshes periodically with interval $\Delta$, and the Segmenter still propagates masks within each segment, but local IDs are no longer associated into video-level global identities. The results show that enabling LIA brings significant HOTA improvements for all five combinations, while Overall HOTA remains at a low level when LIA is disabled. This indicates that segment-level mask propagation alone is insufficient to form video-level global instance trajectories that satisfy the requirements of UAV-OVVIS. By connecting cross-segment local trajectories at the output level, LIA enables scalable segment propagation and long-video global identity maintenance to hold simultaneously.

\begin{figure}[t]
\centering
\includegraphics[width=\columnwidth]{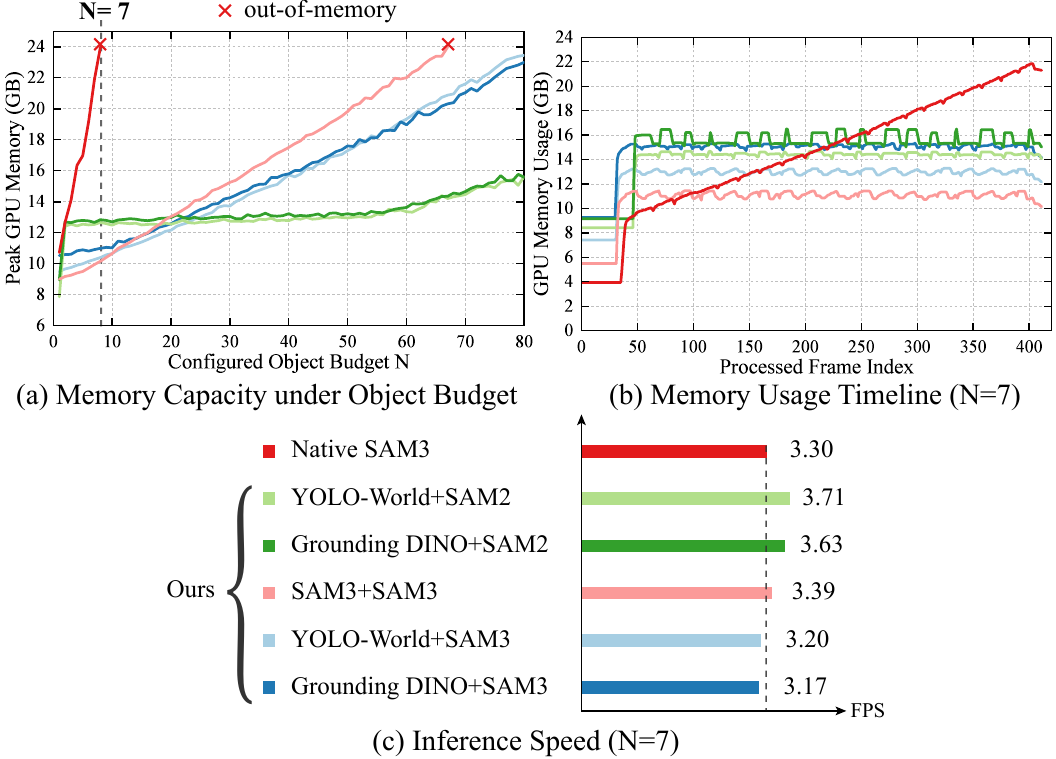}
\caption{Memory capacity, memory trajectory, and inference speed analysis on a stress video. We compare Native SAM3 with our five variants. (a) Peak memory is measured under different target budgets $N$. The vertical dashed line indicates the maximum budget $N=7$ under which Native SAM3 can complete the full run, and $\times$ indicates the first out-of-memory budget. (b) Per-frame memory usage trajectory recorded under $N=7$. (c) Measured FPS under $N=7$.}
\label{fig:memory-stress}
\end{figure}

\subsection{Memory and Inference Speed Analysis}
\label{sec:memory-exp}
To verify the resource advantages of segmented propagation in long-video dense-target scenarios, we further evaluate the memory capacity, long-sequence memory stability, and inference speed of different methods. We select a stress video of approximately 400 frames, where the number of truly visible vehicles remains above 80, and use \textit{car} as the open-vocabulary query. As shown in Figure~\ref{fig:memory-stress}, the memory usage of Native SAM3 continuously increases with the target budget and video length, and the maximum budget under which it can complete the full run is only $N=7$. In contrast, our five variants can run with target budgets more than 10 times larger than that of Native SAM3. The per-frame memory trajectories further show that Native SAM3 continuously accumulates memory in long sequences, whereas AeroTrack constrains the effective memory length through segmented propagation and maintains stable memory usage after the initial loading stage. Meanwhile, the FPS of our five variants is generally comparable to Native SAM3, with some combinations being slightly faster. These results indicate that the segmented design can substantially expand the processable target scale without sacrificing inference speed, making it more suitable for long UAV videos with dense targets.

\section{Conclusion}

This paper introduces UAV-OVVIS, a new task that discovers targets in UAV videos according to open-vocabulary queries and outputs instance-level segmentation trajectories with globally consistent video identities. To address the scarcity of instance-level annotations in UAV scenarios, we propose AeroTrack, a training-free framework that decouples UAV-OVVIS into two cooperative components: a Recognizer and a Segmenter. AeroTrack further maintains cross-segment global identities at the output level through Lifecycle-aware ID Association (LIA), thereby reusing existing visual foundation models to enable open-vocabulary video instance segmentation for UAV videos. We also construct AeroVIS, a dataset and evaluation protocol covering 9 UAV object categories and 8,279 valid trajectories. Experiments show that AeroTrack substantially outperforms existing general OV-VIS methods in the complex UAV scenarios represented by AeroVIS, and demonstrates strong open-vocabulary generalization on three general VIS benchmarks. Further analysis shows that periodic open-vocabulary detection improves target coverage in long videos, segmented propagation substantially expands the processable target budget in dense-target scenarios, and LIA stably maintains cross-segment global identity consistency. Overall, UAV-OVVIS represents a more fine-grained and flexible UAV video perception paradigm, and points to an important direction for UAV perception in open scenarios. Future research may focus on enabling UAV-OVVIS in a more lightweight and efficient manner, while further improving dense small-object perception and global identity maintenance in long sequences.

\FloatBarrier
\bibliography{references}

\end{document}